\documentclass[runningheads,a4paper]{llncs}

\usepackage{amssymb}
\usepackage{graphicx}
\usepackage{textcomp}
\usepackage{soul}  

\usepackage{color}

\usepackage{url}
\newcommand{\keywords}[1]{\par\addvspace\baselineskip
\noindent\keywordname\enspace\ignorespaces#1}

\begin{document}

\mainmatter  

\title{Home Activity Monitoring using Low Resolution Infrared Sensor Array \vspace{-3mm}}

\titlerunning{Home Action Monitoring using Low Resolution Thermal-sensor Array}

%
%
%
\authorrunning{L.Tao et.al}


%
%


\author{Lili Tao\inst{1} \and Timothy Volonakis\inst{1}
Bo Tan \inst{2} \and Yanguo Jing \inst{2}\and \\Kevin Chetty \inst{3} \and Melvyn Smith \inst{1}}

\institute{University of the West of England\\
\email{\{lili.tao, tim.volonakis, melvyn.smith\}@uwe.ac.uk}
\and
Coventry University\\
\email{\{bo.tan, ac2716\}@coventry.ac.uk}
\and
University College London\\
\email{k.chetty@ucl.ac.uk} \vspace{-3mm}}

\maketitle

\begin{abstract} 
Action monitoring in a home environment provides important information for health monitoring and may serve as input into a smart home environment.  Visual analysis using cameras can recognise actions in a complex scene, such as someone’s living room.  However, although there the huge potential benefits and importance, specifically for health, cameras are not widely accepted because of privacy concerns. This paper recognises human activities using a sensor that retains privacy.  The sensor is not only different by being thermal, but it is also of low resolution: 8x8 pixels. The combination of the thermal imaging, and the low spatial resolution ensures the privacy of individuals. We present an approach to recognise daily activities using this sensor based on a discrete cosine transform. We evaluate the proposed method on a state-of-the-art dataset and experimentally confirm that our approach outperforms the baseline method. We also introduce a new dataset, and evaluate the method on it. Here we show that the sensor is considered better at detecting the occurrence of falls and Activities of Daily Living. Our method achieves an overall accuracy of 87.50\% across 7 activities with a fall detection sensitivity of 100\% and specificity of 99.21\%.
\vspace{-3mm}
 \vspace{-3mm}
\end{abstract}

\section{Introduction} \vspace{-3mm}
The U.K., like many other countries, is facing an explosion of long term health conditions. In England alone, 15.4 million people have at least one chronic medical condition, such as, dementia, stroke, cardiovascular or musculoskeletal disease \cite{bajorek2016impact}. In such cases, continuous management and medical treatment may be required for many years outside of hospital. Currently this requires 70\% of the total National Health Services’ (NHS) budget \cite{iacobucci2017nhs}. The huge costs involved could eventually make the NHS unsustainable unless a better solution can be found to reduce costs, while also giving those with long-term conditions better care and an improved quality of life. 

For these reasons, developing a reliable home monitoring system has drawn much attention in recent years due to the growing demands for improved healthcare and patient well-being. Current home monitoring systems often include environmental sensors, wearable inertial sensors and visual sensors. Such systems can enable various types of applications in healthcare provision, such as to help diagnose and manage health and well-being conditions \cite{zhu2015bridging}.

Wearable sensor based techniques have emerged over recent years with a focus on coarse categorisations of activity that offer low cost, low energy consumption, and data simplicity \cite{mukhopadhyay2015wearable}. Among these, tri-axial accelerometers are the most broadly used inertial sensors to recognise ambulation activities \cite{tao2015comparative}. However, despite rapid developments in wearable sensor technology, issues surrounding missed communications, limited battery life, irregular wearing, and poor comfort remain problematic. 

Contactless sensors, such as wireless passive sensing systems and visual sensors, have the potential to address several limitations of the wearable sensors. 
Because of the ubiquitous availability of Wi-Fi signals in indoor areas, wireless signals have been exploited to detect human movement \cite{tan2016awireless} - that is, when a person engages in an activity, the body movement affects the wireless signals. This technology has shown capabilities in assisted living and residential care \cite{tan2018exploiting}. However, Wi-Fi sensing systems still suffer from low accuracy, single-user capability and signal source dependency problems. 
On the other hand, visual sensors can capture rich data and multiple events simultaneously. 
Recent advances in computer vision allow for a fine-grained analysis of human activity that have now opened up the possibility of integrating these devices seamlessly into home monitoring systems \cite{woznowski2015multi}. However, visual sensors have not been widely integrated. This is largely due to the ongoing privacy issue. 

With this in mind, in this paper, we proposed a home activity monitoring system using an 8 $\times$ 8 infrared sensor array.  The sensor provides 64 pixels of thermal data and can be used to offer coarse-grained activity recognition while, importantly, preserving privacy for the users. 
This is a new form of sensor applied in the home monitoring system context, where only limited publicly available datasets are available. We evaluate our method on the \textit{Coventry-Activity} dataset \cite{Karayaneva2018Use}, and compare against a baseline method. We also introduce a new dataset, \textit{Infra-ADL2018}, for monitoring activities of daily living and to detect occurrence of falls. The dataset contains 7 daily activities performed by 8 subjects. The infrared sensor itself is mounted on the ceiling to give an overhead view. In summary, the major contributions of this paper are, (a)Testing the availability of a low resolution thermal sensor to recognise basic human daily activities, (b) exploration of a low resolution thermal sensor in a healthcare scenario to preserve user privacy, and (c)Demonstration that the proposed method is able to achieve high recognition results, especially for detecting fall events. \vspace{-3mm}

\section{Related work} \vspace{-3mm}
Applying computer vision techniques to help with increasing personal safety and reducing risks at home has gained significant attention over recent years. However, studies that address privacy concerns in vision-based home monitoring systems have been relatively limited. Our work therefore explores this field further and builds on several relevant subject areas in vision.

{\textbf{Low resolution infrared sensor - }}
An infrared sensor array is a device composed of a small number of discrete infrared sensors. It represents the spatial distribution of temperature as a low-resolution image. Unlike colour cameras, infrared sensor arrays only capture the shape of the human body, therefore making individual 
identification harder. Additionally, the low spatial resolution also makes identification of individuals difficult.
As this is more comfortable for users, it becomes more acceptable for installation in residential environments. Such infrared sensor arrays can be applied in many scenarios. A 4x4 sensor array has been used to recognise hand motion directions \cite{wojtczuk2011pir}, although the extremely low resolution of this sensor renders it unsuitable for more complex visual tasks. A 8x8 pixel sensor array has been successfully used to detect, count and track people indoors \cite{trofimova2017indoor}. Human movements has also be inferred by using the subject's location and moving trajectory using a 16x16 sensor array \cite{hosono2014human}. Most recently, a multi-sensor system has been designed for human movement detection and activity recognition \cite{Karayaneva2018Use}, which our method will be compared against.

{\textbf{Feature extraction - }}
The visual trace of human activity in video forms a spatio-temporal pattern. Here the salient features are well-developed for images captured by conventional visible-light RGB cameras \cite{aggarwal2011human}. However, the majority of well developed features, such as histogram of oriented gradients or optical flow, are not appropriate and applicable for very low resolution images such as those captured in this study, i.e. for 8x8 pixel resolution images. 

Several features have been investigated specifically for low resolution infrared sensors, most notably \cite{basu2015tracking}.  Here, connected component analysis was used to evaluate the number of individuals present in the scene, which subsequently led to motion tracking of the individuals; however this method was sensitive to background noise.  A thermo-spatial sensitive histogram feature approach was able to reduce the noise from background pixels \cite{hosono2014human}. Although counting and tracking of individuals is a non-trivial task, here we are concerned with the activity of each individual. Intuitively, this would appear to require finer detail, and this poses a difficult task given the low spatial resolution of the image.

{\textbf{Fall detection - }}
A large amount of research is underway in the development of a smart sensing system to detect falls in home environments. However, the use of thermal infrared arrays for fall detection has to date not been widely investigated. Although a real-time system to recognise fall and non-fall events has been presented in \cite{hayashida2017use}, their study overlooks the complexity of non-fall actions, where some actions, such as sitting down and inactivity, can be confused with falling \cite{mubashir2013survey}. Taking this into consideration, various non-fall activities are specifically incorporated in our dataset, including those most likely to be confused with falling. 

\section{Proposed method} 
\begin{figure}[t]
\centering
\includegraphics[scale=0.65]{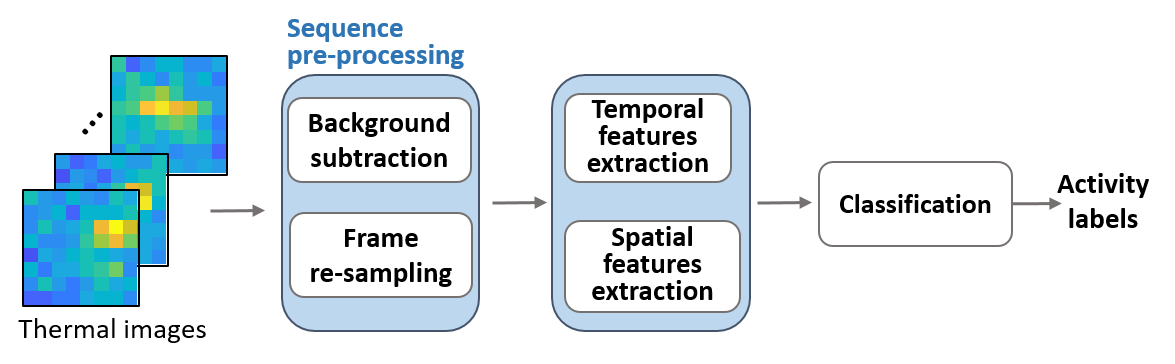} \vspace{-8mm}
\caption{Overview of the proposed framework.}
\label{fig:overview}
\end{figure}

\vspace{-3mm}
We propose a home activity monitoring pipeline to recognise  basic daily activities using thermal images captured by an infrared sensor array. The pipeline of the proposed method is shown in Figure \ref{fig:overview}. Given a set of raw thermal images, a background subtraction is applied first to reduce the affect of noise in background pixels, and then the sequences are re-sampled to the same length. The Discrete Cosine Transform (DCT) based temporal and spatial features are extracted from each sequence. Using these features, the proposed method is able to classify activities and detect falls robustly and accurately.  

\subsection{Sequence pre-processing}
\begin{figure}[t]
\centering
\includegraphics[scale=0.5]{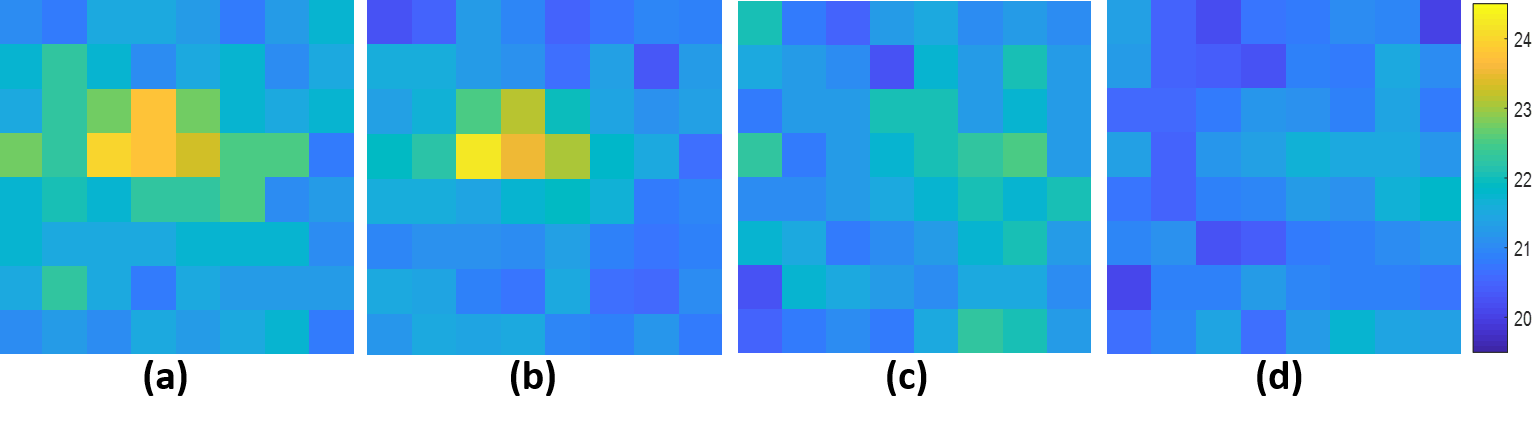} \vspace{-10mm}
\caption{Background subtraction examples. (a) raw image with a subject; (b) background subtracted image with a subject; (c) raw image without a subject and (d) background subtracted image without a subject. }
\label{fig:BS}
\end{figure}

The raw data collected by the infrared sensor array is noisy, especially in the background pixels. A sequence of background scene images with $F$ frames is taken without human subjects in it. The background image $B_i$ is formed by averaging of all $i^{th}$ pixels along the sequence $B_i = \frac{1}{F}\sum_{f=1}^{F}{B_{i}^f} $. The processed background subtracted image is extracted by subtracting the corresponding pixel at the background image, $P_{i} = \hat{P}_{i} - B_i$, where $P_i$ and $\hat{P}_i$ are the processed image and raw image, respectively. 
Figure \ref{fig:BS} { shows examples of the background subtraction with and without human subjects present in the image }. Subtracting the average background pronounces the spatial energy of an individual when the individual is present and inhibits energy in the image when an individual is absent from the scene.

{ The length of the sequences are different. This is because the duration of some actions are longer than others, e.g. sitting down is often a faster action to complete than walking across a room.  It may have been possible to design the data collection to reduce this problem for analysis; perhaps to have a fixed time to record all actions, or to process the data differently i.e. crop all actions to be of the same length as the fastest action. This however, although important, is outside the scope of this paper; instead we are presenting a home monitoring approach using a low resolution sensor. The data has been pre-processed so that each action has an equal number of frames.  This was achieved by sampling at equal intervals.  }

\subsection{Temporal and spatial feature extraction}
First, a one-dimensional DCT is performed on the time signals of a series of images, and the temporal feature vector is created by the DCT coefficients. The advantage of using the DCT is the ability to compactly represent an activity sequence using a fixed number of coefficients. Then, a two-dimensional DCT is performed for each image. 

A temporal domain feature is used to represent a time series $\mathbf{P}_i = \{P^1_i, P^2_i,..., P^F_i\}$ for $i^{th}$ pixel in the frequency domain by the 1-D DCT, where the feature takes the absolute value of the $k$ lowest frequency components of the frequency coefficients, \vspace{-2mm}
\begin{equation}
\mathcal{F}_{1D}(\mathbf{P}_i) = |X_{1:k}\mathbf{P}_i|
\end{equation}
where $X$ is the discrete cosine transformation matrix. The temporal feature for all $N$ pixels can be written as $\{\mathcal{F}_{1D}(\mathbf{P}_1), \mathcal{F}_{1D}(\mathbf{P}_2),...,\mathcal{F}_{1D}(\mathbf{P}_N)\}$. For our experiments, we use the first 5 lowest frequency components of the frequency coefficients.

A two-dimensional DCT for each image is calculated to form spatial domain features. This results in a matrix of 8 $\times$ 8 coefficients. A subset of these values is taken to construct the feature vector, where the low-frequency components within the processed image are chosen. Similar to a 1-D DCT, the spatial DCT feature for each frame is,  \vspace{-3mm}
\begin{equation}
\mathcal{F}_{2D}(\mathbf{P}^f) = |Y_{1:k,1:k}\mathbf{P}^f|
\end{equation}
where $Y$ is the 2-D discrete cosine transformation matrix. The spatial feature for all $F$ frames can be written as $\{\mathcal{F}_{2D}(\mathbf{P}^1), \mathcal{F}_{2D}(\mathbf{P}^2),...,\mathcal{F}_{2D}(\mathbf{P}^F)\}$. In our experiment, we use a set of 3 $\times$ 3 coefficients located in the upper left corner of the coefficients matrix. 

The activity is then inferred via a multi-class linear Support Vector Machine using the concatenation of the temporal and spatial features. 

\section{Experimental results}
In our experiments, all the data are collected from the Grid-EYE 8 $\times$ 8 infrared sensor array developed by Panasonic \cite{GridEye}.
\subsection{Datasets}
The \textbf{Coventry-Activity} dataset \cite{Karayaneva2018Use} is designed for evaluation of activity recognition under a multi-sensor setting. The three sensors are placed 1.5 meters away from the subject as follows: in front, to the left and to the right. 8 activities are collected with one subject in the scene. The dataset contains 3 subjects performing each activity 10 times. We evaluate our activity recognition method on this dataset and compare against the baseline method \cite{Karayaneva2018Use}.\\
The \textbf{Infra-ADL2018} dataset is introduced in this paper for monitoring home activities and detect occurrence of falls. The dataset is generated over 24 sessions by 8 subjects containing 7 activity categories per session: \textit{fall, sit still, stand still, sit to stand, stand to sit, walking from left to right}, and \textit{walking from right to left}.

\subsection{Quantitative evaluation}
\begin{figure}[t]
\centering
\includegraphics[scale=0.3]{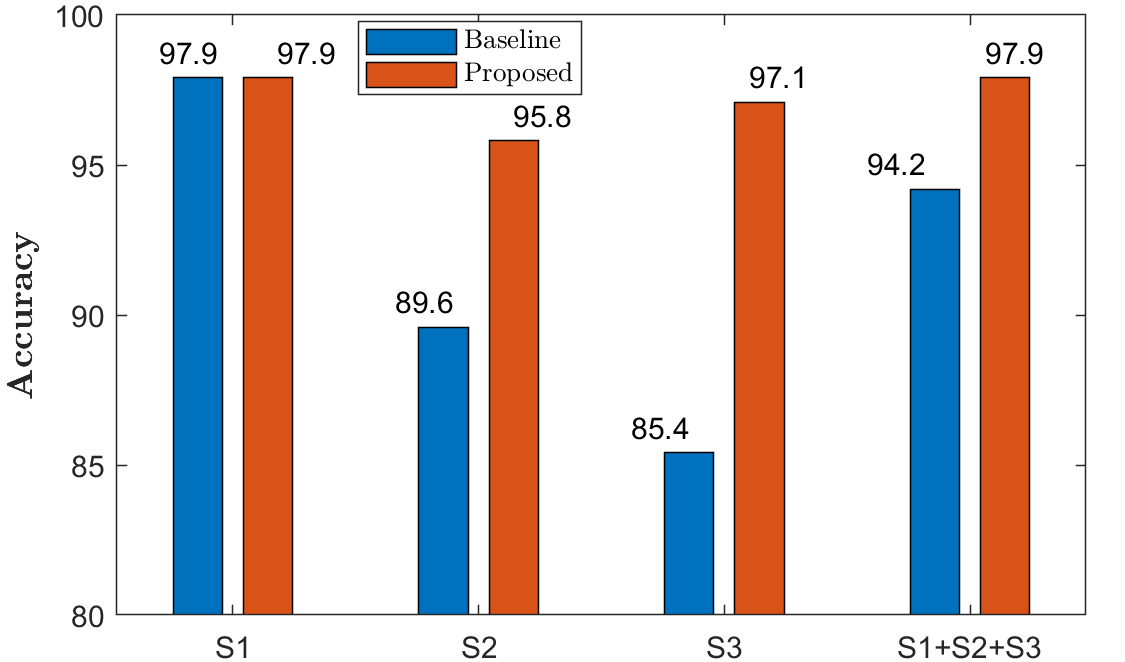}
\caption{Average recognition accuracy by using different sensors for the proposed method and the baseline method.}
\label{fig:compare}
\end{figure}

We first compare the proposed method against the baseline method presented in \cite{Karayaneva2018Use} on the \textit{Coventry-Activity} dataset. To follow the same setting, we perform the 10-fold cross validation to the dataset, and test using three sensors solely and also when fusing them together. Sensor 1(S1), Sensor 2 (S2) and Sensor 3 (S3) are placed on the right side, in front and on left side of the subject, respectively. The results are shown in Figure \ref{fig:compare} where it can be seen that the proposed method significantly outperforms the baseline when S2 and S3 are used solely, and also when the three sensors are used together. Table \ref{tab:compare} shows the average recognition accuracy for each activity when the three sensors are used together - with the best results for each activity highlighted. We note that the proposed method achieves high recognition accuracy throughout all activities. Unlike the baseline method which produces very low accuracy for some activities, e.g. 50\% for move left to right and 57\% for move forward and backward. 

\begin{table}[t]
\begin{tabular}{|c|p{1cm}|p{1cm}|p{1.8cm}|p{1cm}|p{1.5cm}|p{1.9cm}|p{1.25cm}|p{1.25cm}|}
\hline 
 & sit still & stand still & stand up and sit down  & stand up & move left to right & move forward to backward & diagonal walk 1  & diagonal walk 2\tabularnewline
\hline 
\hline 
Baseline & \textbf{100} & \textbf{100} & 86 & \textbf{100} &50  & 57 & 67 &\textbf{100} \tabularnewline
\hline 
Proposed & 96 & 93 & \textbf{96}& \textbf{100} &  \textbf{100}&\textbf{96}  & \textbf{100} &\textbf{100} \tabularnewline
\hline 
\end{tabular}
\caption{The average recognition accuracy for each activity when three sensors are used together. The best results in each activity are in bold.}
\label{tab:compare}
\end{table}

\begin{figure}[t]
\centering
\includegraphics[scale=0.3]{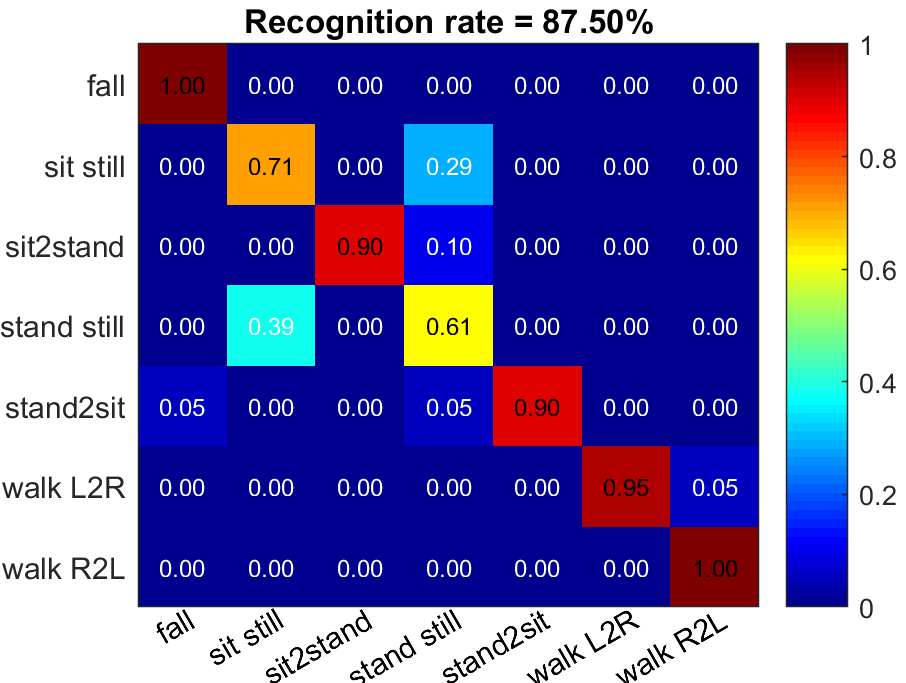}
\caption{The confusion matrix of the activity recognition for the \textit{Infra-ADL2018} dataset, corresponding to an average recognition rate of 87.50\%}
\label{fig:confusionMatrix}
\end{figure}

We then test the proposed method on our new dataset \textit{Infra-ADL2018}. We perform a leave-one-subject-out cross validation where final recognition results reported are averaged over all subjects to remove any bias. The confusion matrix of the activity recognition is shown in Figure \ref{fig:confusionMatrix}. In general, the overall recognition rate of the proposed method is 87.50\%. More precisely, it can reach 100\% sensitivity for detecting the occurrence of fall, and 99.21\% specificity, indicating very few false fall alarms. The only action that is misclassified as a fall is stand to sit. This is likely to be because some subjects perform the action very fast, in which case the action looks similar to a fall. The most confused actions are the sit still and stand still due to the sensor position of the system. The sensor is mounted at on the ceiling so that the sit still and stand still look very similar in the thermal images, where the same few pixels are shown higher in temperature throughout the sequence. The use of a multi-sensor setting is one way to discriminate between these by recognising the shape of human when performing different actions.

\section{Conclusion and Future work} \vspace{-3mm}
In this paper, we proposed a home activity monitoring method using an infrared sensor array. The proposed method uses a temporal and spatial Discrete Cosine Transform that is suitable for representing human activity in very low-resolution thermal images. Given that the sensor would be used in home environments, potential future directions include a multi-sensor system that comprises multiple viewing angles that can deal with view-invariance and occlusion.

{
\bibliographystyle{abbrv}
\bibliography{workshop}
}

\end{document}